\documentclass[a4paper,twoside]{article}
\usepackage{epsfig}
\usepackage{subcaption}
\usepackage{calc}
\usepackage{amssymb}
\usepackage{amstext}
\usepackage{amsmath}
\usepackage{amsthm}
\usepackage{multicol}
\usepackage{apalike}
\usepackage{SCITEPRESS}
\usepackage{graphicx}
\usepackage{xcolor}
\usepackage{amsmath}
\usepackage{hyperref}
\usepackage{url}
\usepackage{mathtools}
\usepackage{commath}
\usepackage[ruled,vlined]{algorithm2e}
\usepackage{siunitx}
\usepackage{textcase}
\begin{document}
\title{Continuous Perception for Classifying Shapes and Weights of Garments for Robotic Vision Applications}

\author{\authorname{Li Duan\sup{1}\orcidAuthor{0000-0002-0388-752X
}, Gerardo Aragon-Camarasa\sup{1}\orcidAuthor{0000-0003-3756-5569}}
\affiliation{\sup{1}School of Computing Science, University of Glasgow, Glasgow, United Kingdom}
\email{l.duan.1@research.gla.ac.uk, gerardo.aragoncamarasa@glasgow.ac.uk}
}

\keywords{\centering Continuous Perception, Depth Images, Shapes and Weights Predictions} 

\abstract{We present an approach to continuous perception for robotic laundry tasks. Our assumption is that the visual prediction of a garment's shapes and weights is possible via a neural network that learns the dynamic changes of garments from video sequences. Continuous perception is leveraged during training by inputting consecutive frames, of which the network learns how a garment deforms. To evaluate our hypothesis, we captured a dataset of 40K RGB and depth video sequences while a garment is being manipulated. We also conducted ablation studies to understand whether the neural network learns the physical properties of garments. Our findings suggest that a modified AlexNet-LSTM architecture has the best classification performance for the garment's shapes and discretised weights. To further provide evidence for continuous perception, we evaluated our network on unseen video sequences and computed the 'Moving Average' over a sequence of predictions. We found that our network has a classification accuracy of 48\% and 60\% for shapes and weights of garments, respectively.}

\onecolumn \maketitle \normalsize \setcounter{footnote}{0} \vfill

\section{\uppercase{INTRODUCTION}}

Perception and manipulation in robotics are an interactive process which a robot uses to complete a task \cite{8007233}. That is, perception informs manipulation, while manipulation of objects improves the visual understanding of the object. Interactive perception predicates that a robot understands the contents of a scene visually, then acts upon it, i.e. manipulation starts after perception is completed. In this paper, we depart from the idea of interactive perception and theorise that perception and manipulation run concurrently while executing a task, i.e. the robot perceives the scene and updates the manipulation task continuously (i.e. continuous perception). We demonstrate continuous perception in a deformable object visual task where a robot needs to understand how objects deform over time to learn its physical properties and predict the garment's shape and weight.

Due to their materials’ physical and geometric properties, garments usually have folds, crumples and holes, which are irregular shaped and configured, making garments to have a high-dimensional state space. Therefore, when a robot manipulates a garment, the deformations of the garment is unpredictable and complex. Due to the high dimensionality of garments and complexity in scenarios while manipulating garments, previous approaches for predicting categories and physical properties of garments are not robust to continuous deformations \cite{tanaka2019learning}\cite{MARTINEZ2019220}. Prior research \cite{bhat2003estimating}\cite{tanaka2019learning}, \cite{runia2020cloth} has leveraged the use of simulated environments to predict how a garment deforms, however, real-world manipulation scenarios such as grasping, folding and flipping garments are difficult to be simulated because garments can take an infinite number of possible configurations in which a simulation engine may fail to capture. Moreover, simulated environments can not be fully aligned with the real environment, and a slight perturbation in the real environment will cause simulations to fail. 

Garment configurations (garment state space) have high dimensionality, therefore motion planning for garments requires a high dimensionality space. This motion planning space should represent the dynamic characteristics of garments and the robot's dynamic capabilities to successfully plan a motion. In this paper, we argue that it is need to learning physical properties first to enable robotic manipulation of garments (and other deformable objects). Garment shapes and weights are two important geometric and physical properties of garments. In this paper, we therefore learn the physical and geometric properties of garments from real-world garment samples. For this, garments are being grasped from the ground and then dropped. This simple manipulation scenario allows us  to  train a  neural  network to  perceive dynamic changes  from  depth images and  learn physical (weights) and geometric (shapes) properties of garments while being manipulated, see Fig \ref{fig:OverallArchitecture}. 

To investigate the continuous perception of deformable objects, we have captured a dataset containing video sequences of RGB and depth images. We aim to predict the physical properties (i.e. weights) and categories of garment shapes from a video sequence. Therefore, we address the state-of-the-art limitations by learning dynamic changes as opposed to static representations of garments \cite{jimenez2020perception,ganapathi2020learning}. We use weight and shape as the experimental variables to support our continuous perception hypothesis. We must note that we do not address manipulation in this paper since we aim to understand how to equip a robot best to perceive deformable objects visually, as serves as a prerequisite for accommodating online feedback corrections for garment robotic manipulation. Our codes and datasets are available at: \url{https://github.com/cvas-ug/cp-dynamics}

\section{\uppercase{BACKGROUND}}

Bhat \cite{bhat2003estimating} proposed an approach to learn physical properties of clothes from videos by minimising a squared distance error (SSD) between the angle maps of folds and silhouettes of the simulated clothes and the real clothes. However, their approach observes high variability while predicting physical properties of clothes such as shear damping, bend damping and linear drag. Li \textit{et al.}  \cite{li2018learning,li2020visual} has proposed to integrate particles to simulate simple fabrics and fluids in order to learn rigidness and moving trajectories of a deformable object using a Visually Grounded Physics Learner network (VGPL). By leveraging VGPL together with an LSTM, the authors can predict the rigidness and future shapes of the object. In their research, they are using particles to learn the dynamic changes of objects. In contrast, due to the high dimensionality and complexity of garments, particles are an approximation to the dynamic changes which cannot be fully described for a robot manipulation task. In Bhat\cite{bhat2003estimating}’s work, they study the physical properties of garments by comparing differences between motion video frames of real and simulated garments, where they only aim to match the shapes between real and simulated garments. In our work, we learn the physical properties of garments through continuously perceiving video frames of grasped garments, which means we learn deformations of garments rather than shapes of garments. Deformations of garments encode dynamics characteristics of garments, which depend on their physical and geometric properties. Therefore, our network has higher accuracy while predicting physical and geometric properties of garments (equivalent to shapes and weights of garments) compared to \cite{bhat2003estimating}.

To learn elasticity of objects, Senguapa \textit{et al.} \cite{sengupta2020simultaneous} has proposed an approach where a robot presses the surface of objects and observes the object's shape changes in a simulated and a real environment. They aimed to find the difference of the simulated and real objects Young's modules to estimate the object's elasticity and estimate forces applied on the object without any force sensor. Tanake \textit{et al.} \cite{tanaka2019learning} minimised the shape difference between real and simulated garments to find their stiffness. In these two approaches, if there exists a small variation between simulation and reality or if an unseen object is presented, their approaches require to simulate novel object models again as the simulation is limited to known object models.

Compared with previous research that has utilised temporal images to analyse the physical properties of deformable objects, Davis \textit{et al.} \cite{davis2015visual} chose to investigate deformable objects' physical properties in terms of their vibration frequencies. That is, they employed a loudspeaker to generate sonic waves on fabrics to obtain modes of vibration of fabrics and analysed the characteristics of these modes of vibration to identify the fabrics materials. The main limitation of this approach is in the use of high-end sensing equipment which would make it impractical for a robotic application. In this paper, we employ an off-the-shelf RGBD camera to learn dynamic changes of garments.

Yang \textit{et al.} \cite{yang2017learning} has proposed a CNN-LSTM architecture. Their method consists of training a CNN-LSTM model to learn the stretch stiffness and bend stiffness of different materials and then apply the trained model to classify garment material types. However, suppose a garment consists of multiple materials. In that case, the CNN-LSTM model will not be able to predict its physical properties because their work focuses on garments with only one fabric type. Mariolis \textit{et al.} \cite{mariolis2015pose} devised a hierarchical convolutional neural network to conduct a similar experiment to predict the categories of garments and estimate their poses with real and simulated depth images. Their work has pushed the accuracy of the classification from 79.3\% to 89.38\% with respect to the state of the art. However, the main limitations are that their dataset consists of 13 garments belonging to three categories. In this paper, we address this limitation by compiling a dataset of 20 garments belonging to five categories of similar material types, and we have evaluated our neural network to predict unseen garments.

Similar to this work, Martinez \textit{et al.} \cite{MARTINEZ2019220} has proposed a continuous perception approach to predict the categories of garments by extracting Locality Constrained Group Sparse representations (LGSR) from depth images of the garments. However, the authors did not address the need to understand how garments deform over time continuously as full sequences need to be processed in order to get a prediction of the garment shape. Continuous predictions is a prerequisite for accommodating dexterous robotic manipulation and online feedback corrections to advanced garment robotic manipulation.

\section{\uppercase{MATERIALS and METHODS} \label{sec:matandmethods}}

\begin{figure*}[t]
    \centering
    \includegraphics[width=0.8\textwidth]{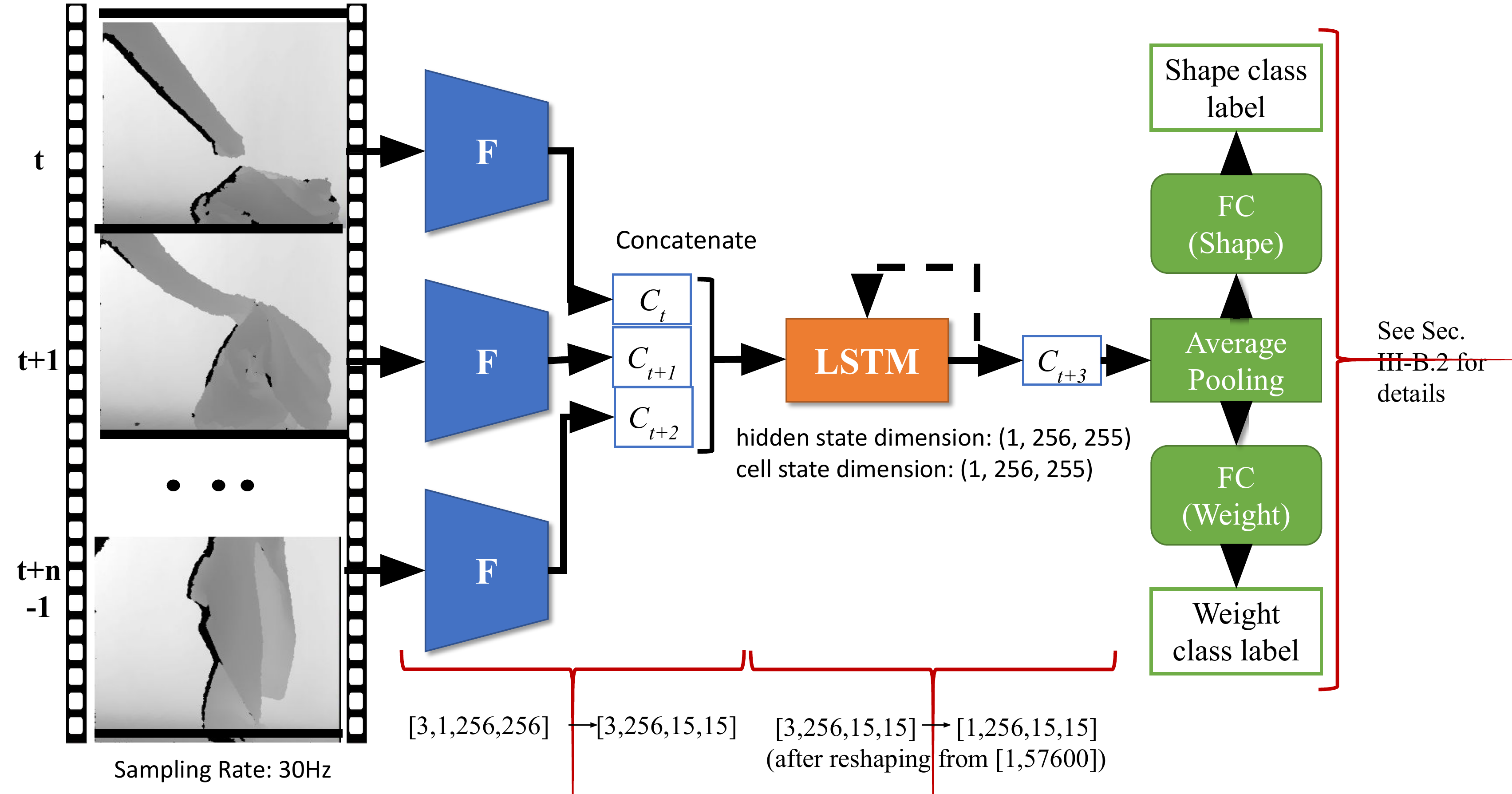}
    \caption{Our network is divided into feature extraction (F), an LSTM unit and classifier networks. Depth images of a garment with a resolution of $256\times256$ pixels are passed to the feature extraction network. Three feature latent spaces, i.e. $C_t$, $C_t+1$ and $C_t+2$ from time-steps $t$, $t+1$ and $t+2$, respectively, are concatenated and then passed to the LSTM. Each feature latent space has a tensor size of $15\times15$ with a channel size of 256. From the LSTM, we obtain a predicted future feature latent space ($C_t+3$) which is reshaped back to the original feature space size (i.e. $[1,256,15,15]$) and input to an average pooling layer. The average pool output with size of $[1,256,6,6]$ is flattened to $[1,9216]$ and passed to the fully connected (FC) shape and weight classifiers.}
    \label{fig:OverallArchitecture}
\end{figure*}

We hypothesise that \textit{continuous perception allows a robot to learn the physical properties of clothing items implicitly (such as stiffness, bending, etc.) via a Deep Neural Network (DNN) because a DNN can predict the dynamic changes of an unseen clothing item above chance}. For this, we implemented an artificial neural network that classifies shapes and weights of unseen garments (Fig. \ref{fig:OverallArchitecture} and Section \ref{sec:netdesign}). Our network consists of a feature extraction network, an LSTM unit and two classifiers for classifying the shape and weight of garments. We input three consecutive frame images $(t,t+1,t+2)$ into our network to predict the shape and weight of the observed garment from a predicted feature latent space at $t+3$. We propose to use the garment's weight as an indicator that the network has captured and can interpret the physical properties of garments. Specifically, the garment's weight is a physical property and is directly proportional to the forces applied to the garment's fabric over the influence of gravity.

\paragraph{Garment Dataset \label{sec:dataset}} To test our hypothesis, we have captured 200 videos of a garment being grasped from the ground to a random point above the ground around 50 cm and then dropped from this point. Each garment has been grasped and dropped down ten times in order to capture its intrinsic dynamic properties. Videos were captured with an ASUS Xtion Pro, and each video consists of 200 frames (the sampling rate is 30Hz), resulting in 40K RGB and 40K depth images at a resolution of 480$\times$680 pixels.

Our dataset features 20 different garments of five garment shape categories: pants, shirts, sweaters, towels and t-shirts. Each shape category contains four unique garments. Garments are made of cotton except for sweaters which are made of acrylic and nylon. To obtain segmentation masks, we use a green background, and we used a green sweater to remove the influence of our arm.
We then converted RGB images to a $HSV$ colour space and identified an optimal thresholding value in the $V$ component to segment the green background and our arm from the garment.

\subsection{Network Architecture\label{sec:netdesign}}

Our ultimate objective is to learn the dynamic properties of garments as they are being manipulated. For this, we implemented a neural network comprising a feature extraction network, a recurrent neural network, and a shape and a weight classifier networks. Fig. \ref{fig:OverallArchitecture} depicts the overall neural network architecture. We split training this architecture into learning the appearance of the garment in terms of its shape first, then learning the garments dynamic properties from visual features using a recurrent neural network (i.e. LSTM).

\paragraph{Feature extraction\label{sec:fe}}

A feature extraction network is needed to describe the visual properties of garments (RGB images) or to describe the topology of garments (depth images). We therefore implemented 3 state of the art network architectures, namely AlexNet\cite{krizhevsky2012imagenet},VGG 16\cite{simonyan2014very} and ResNet 18 \cite{he2016deep}. In Section \ref{sec:continuous}, we evaluate their potential for extraction features from garments.

\paragraph{Shape and weight classifiers\label{sec:classifier}}
The classifier components in AlexNet, Resnet and VGG-16 networks comprise fully connected layers that are used to predict a class depending on the visual task. In these layers, one fully connected layer is followed by a rectifier and a regulariser, i.e. a ReLu and dropout layers. However, in this paper, we consider whether the dropout layer will benefit the ability of the neural network to generalise the classification prediction for garments. The reason is that the image dataset used to train these networks contain more than 1000 categories and millions of images \cite{krizhevsky2012imagenet}, while our dataset is considerable smaller (ref. Section \ref{sec:dataset}). The latter means that the dropout layers may filter out useful features while using our dataset. Dropout layers are useful when training large datasets to avoid overfitting. Therefore, we have experimented with modifying the fully connected networks by removing the ReLu and dropout layers and observe their impact on the shape and weight classification tasks. After experimenting with four different network parameters, we found that the best performing structure comprises three fully connected layer blocks, each of which only contains a linear layer. The number of features stays as 9216 without any reduction, then the number reduces to 512 in the second layer, and finally, we reduce to 5 for shape and 3, for weight as the outputs of the classifications. We do not include these experiments in this paper as they do not directly test the hypothesis of this paper but instead demonstrates how to optimise the classification networks for the shape and weight classifiers in this paper.

\paragraph{LSTM Rationale\label{sec:lstm}}
The ability to learn dynamic changes of garments is linked to perceiving the object continuously and being able to predict future states. That is, if a robot can predict future changes of garments, it will be able to update a manipulation task on-the-fly by perceiving a batch of consecutive images rather then receiving a single image and acting sequentially. For this, we have adopted a Long Short-Term Memory (LSTM) network to learn the dynamic changes of consecutive images. After training (ref. Section \ref{sec:training}), we examined the ability to learn garments' dynamic changes by inputting unseen garments images into the trained feature extractor to get their encoded feature maps and input those encoded feature maps into the trained LSTM and evaluate if the network (Fig. \ref{fig:OverallArchitecture}) can predict shapes and weights classifications.

\subsection{Training Strategy \label{sec:training}}

We split training our architecture (Fig. \ref{fig:OverallArchitecture}) into two parts. First, we let the network learn the appearance or topology of garments by means of the feature extraction and classification networks (Sections \ref{sec:fe} and \ref{sec:classifier}). After this, we then train the LSTM network while freezing the parameters of the feature extraction and classification networks to learn the dynamic changes of garments.

We have used pre-trained architectures for AlexNet, Resnet 18 and VGG 16 but fine-tuned its classifier component. For depth images, we fine-tuned the input channel size of the first convolutional layer from 3 to 1 (for AlexNet, Resnet 18 and VGG 16). The loss function adopted is Cross-Entropy between the predicted shape label and the target shape label. After training the feature extraction networks, we use these networks to extract features of consecutive images and concatenate features for the LSTM. The LSTM learning task is to predict the next feature description from the input image sequence, and this predicted feature description is passed to the trained classifier to obtain a predicted shape or weight label. The loss function for training the LSTM consists of the mean square error between the target feature vector and the predicted feature vector generated by the LSTM, and the Cross-Entropy between the predicted shape label and the target shape label. The loss function is:

\begin{equation}
\mathcal{L}_{total}=\mathcal{L}_{MSE}+1000\times\mathcal{L}_{Cross-Entropy} 
\end{equation}

We have used a 'sum' mean squared error during training, but we have reported our results using the average value of the mean squared error of each point in the feature space. We must note that we multiply the cross-entropy loss by 1000 
to balance the influence of the mean squared error and cross-entropy losses.

\section{\uppercase{EXPERIMENTS}\label{sec:experiments}}

For a piece of garment, shape is not an indicator of the garment's physical properties but the garment's weight as it is linked to the material's properties such as stiffness, damping, to name a few. However, obtaining ground truth for stiffness, damping, etc. requires the use of specialised equipment and the goal of this paper is to learn these physical properties implicitly. That is, we propose to use the garment's weight as a performance measure to validate our approach using unseen samples of garments.

To test our hypothesis, we have adopted a leave-one-out cross-validation approach. That is, in our dataset, there are five shapes of garments: pants, shirts, sweaters, towels and t-shirts; and for each type, there are four garments (e.g. shirt-1, shirt-2, shirt-3 and shirt-4). Three of the four garments (shirt-1, shirt-2 and shirt-3) are used to train the neural network, and the other (shirt-4) is used to test the neural work (unseen samples). We must note that each garment has different appearance such as different colour, dimensions, weights and volumes. For weight classification, we divided our garments into three categories: light (the garments weighed less than 180g), medium (the garments weighed between 180g and 300g) and heavy (the garments weighted more than 300g).

We have used a Thinkpad Carbon 6th Generation (CPU: Intel i7-8550U) equipped with an Nvidia GTX 970, running Ubuntu 18.04. We used \textit{SGD} as the optimiser for training the feature extraction and classification networks, with a learning rate of \num{1e-3} and a momentum of 0.9 for $35$ epochs. We then used \textit{Adam} for training the LSTM with a learning rate of \num{1e-4} and a step learning scheduler with a step size of $15$ and decay rate of $0.1$ for $35$ epochs. The reason for adopting different optimisers is that Adam provides a better training result than SGD for training the LSTM, while SGD observes faster training for the feature extraction and classifiers. To test our hypothesis, we first experiment on which image representation (RGB or depth images) is the best to capture intrinsic dynamic properties of garments. We also examined three different feature extraction networks to find the best performing network for classifying shapes and weights of garments (Section \ref{sec:exp-fe}). 
Finally, we evaluate the performance of our network on a continuous perception task (Section \ref{sec:continuous}).

\begin{table*}[t]
    \caption{Classification accuracy (in percentages) of unseen garment shapes}
    \label{tab:comparison}
    \centering
    \begin{tabular}{|c|c|c|c|c|c|c|}
    \hline
     Feature Extractor     &  P & SH & SW & TW & TS & \textit{\textbf{Average}}\\
     \hline
     AlexNet\textbf{(depth)} &  57.0 & 13.0 & 71.0 & 47.0 & 50.0 & \textit{\textbf{47.6}}\\
     \hline
     AlexNet\textbf{(RGB)} & 18.0 & 13.0 &0.0 & 0.0 & 0.0 & \textit{\textbf{6.2}}\\
     \hline
     VGG16\textbf{(depth)} & 25.0 & 18.0 & 35.0 & 20.0 & 25.0 & \textit{\textbf{24.6}}\\
     \hline
     VGG16\textbf{(RGB)} & 9.0 & 14.0 & 20.0 & 7.0 & 11.0 & \textit{\textbf{12.2}}\\
     \hline
     ResNet18\textbf{(depth)} & 6.0 & 10.0 & 51.0 & 69.0 & 5.0 & \textit{\textbf{28.2}} \\
     \hline
     ResNet18\textbf{(RGB)} & 16.0  & 1.0 & 2.0 & 81.0 & 14.0 & \textit{\textbf{22.8}} \\
     \hline
    \end{tabular}
\end{table*}


\begin{table*}[t]
    \caption{Classification accuracy of unseen garment weights.}
    \label{tab:weight-comparison}
    \centering
    \begin{tabular}{|c|c|c|c|c|}
    \hline
     Feature Extractor     &  Light & Medium & Heavy &\textit{\textbf{Average}}\\
     \hline
     AlexNet \textbf{(depth)}&  72.0 & 18.0 & 55.0 & \textit{\textbf{48.3}}\\
     \hline
     AlexNet \textbf{(RGB)}&  82.0 & 14.0 & 7.0 & \textit{\textbf{34.3}}\\
     \hline
     VGG16 \textbf{(depth)}& 40.0 & 48.0 & 31.0 & \textit{\textbf{39.7}}\\
     \hline
     VGG16 \textbf{(RGB)}& 38.0 & 3.0 & 100.0 & \textit{\textbf{47}} \\
     \hline
     ResNet18 \textbf{(depth)}& 51.0 & 6.0 & 47.0 & \textit{\textbf{34.7}} \\
     \hline
     ResNet18 \textbf{(RGB)}& 41.0 & 5.0 & 10.0 & \textit{\textbf{18.7}} \\
     \hline
    \end{tabular}
\end{table*}


\subsection{Feature Extraction Ablation\label{sec:exp-fe}}

We have tested using three different deep convolutional feature extraction architectures: AlexNet, VGG 16 and ResNet 18. We compared the performance of shape and weight classification of unseen garments with RGB and depth images. These feature extractors have been coupled with a classifier without an LSTM; effectively producing single frame predictions similar to \cite{sun2017singleshot}.

From Table \ref{tab:comparison}, it can be seen that ResNet 18 and VGG 16 overfitted the training dataset. As a consequence, their classification performance is below or close to a random prediction, i.e. we have 5 and 3 classes for shape and weight. AlexNet, however, observes a classification performance above chance for depth images. By comparing classification performances between RGB and depth images in Table \ref{tab:comparison}, we observe that depth images (47.6\%) outperformed the accuracy of a network trained on RGB images (7.4\%) while using AlexNet. The reason is that a depth image is a map that reflects the distances between each pixel and the camera, which can capture the topology of the garment. The latter is similar to the findings in \cite{sun2017singleshot,MARTINEZ2019220}.

We observe a similar performance while classifying garments' weights. AlexNet has a classification performance of $48.3\%$ while using depth images. We must note that the weights of garments that are labelled as 'medium' are mistakenly classified as 'heavy' or 'light'. Therefore compared to the predictions on shape, predicting weights is more difficult for our neural network on a single shot perception paradigm. From these experiments, we, therefore, choose AlexNet as the feature extraction network for the remainder of the following experiments.

\subsection{Continuous Perception\label{sec:continuous}}

To test our continuous perception hypothesis (Section \ref{sec:matandmethods}), we have chosen AlexNet and a window sequence size of 3 to predict the shape and weight of unseen video sequences from our dataset, i.e. video sequences that have not been used for training. For this, we accumulate prediction results over the video sequence and compute the Moving Average (MA) over the evaluated sequence. That is, MA serves as the decision-making mechanism that determines the shape and weight classes after observing a garment deform over time rather than the previous three frames as in previous sections. 

\begin{figure}[t]
    \centering
    \includegraphics[width=0.45\textwidth]{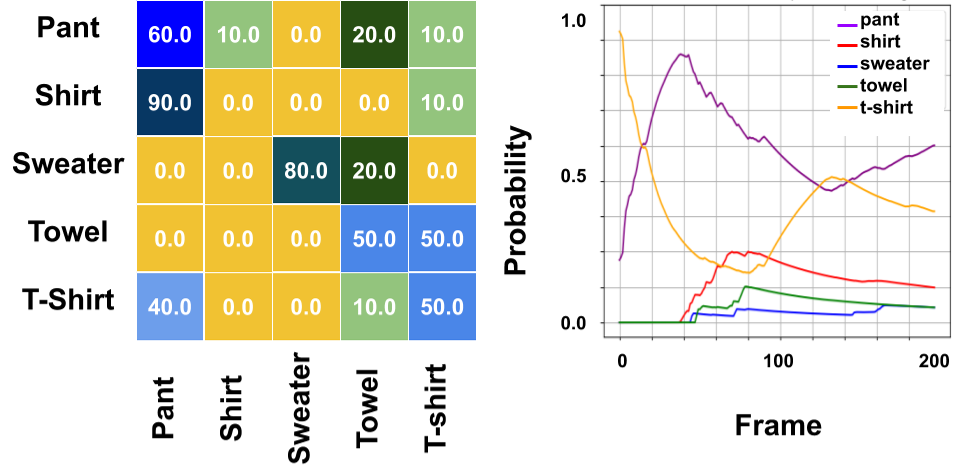}
    \caption{Continuous shape prediction (Left: \textit{Moving Average Confusion Matrix}; Right: \textit{Moving Average over a video sequence})}
    \label{fig:continuous}
\end{figure}

\begin{figure}[t]
    \centering
    \includegraphics[width=0.45\textwidth]{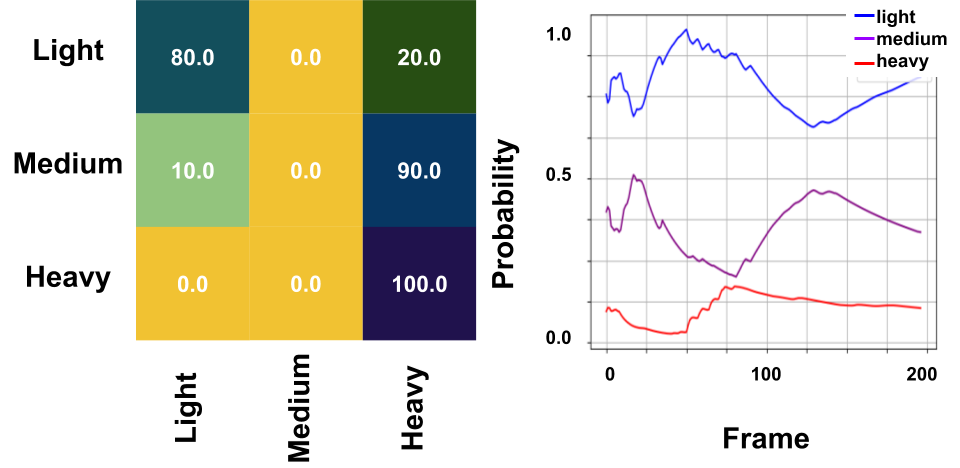}
    \caption{Continuous weight prediction (Left: \textit{Moving Average Confusion Matrix}; Right: \textit{Moving Average over a video sequence})}
    \label{fig:sequence}
\end{figure}

This experiment consists of passing 3 consecutive frames to the network to output a shape and weight class probability for each output in the networks. We then compute their MA values for each output before sliding into the next three consecutive frames, e.g. slide from frame $t-2$, $t-1$, $t$ to frame $t-1$, $t$, $t+1$. After we slide across the video sequence and accumulate MA values, we calculated an average of the MA values for each class. We chose the class that observes the maximum MA value as a prediction of the target category. Our unseen test set contains 50 video sequences. Hence, we got 50 shape and weight predictions and used to calculate the confusion matrices in Fig. \ref{fig:continuous} and Fig. \ref{fig:sequence}.

From Fig. \ref{fig:continuous}(\textit{left}) and Fig. \ref{fig:sequence}(\textit{left}), it can been seen that an average prediction accuracy of $48\%$ for shapes and an average prediction of $60\%$ for weights have been obtained for all unseen video sequences. We can observe in Fig. \ref{fig:continuous}(\textit{left}) that the shirt has been wrongly classified as a pant in all its video sequences, but the sweater is labelled correctly in most of its sequences. Half of the towels have been wrongly recognised as a t-shirt. Also for weight, the medium-weighted garments are wrongly classified in all their sequences, where most of them have been categorised as heavy garments, but all heavy garments are correctly classified. Fig. \ref{fig:continuous} (\textit{right}) shows an example of the MA over a video sequence of a shirt. It can be seen that the network changes its prediction between being a t-shirt or a pant while the correct class is a shirt. The reason for this is that the shirts, t-shirts and pants in our dataset are made of cotton. Therefore, these garments have similar physical properties, but different shapes and our neural network is not capable of differentiating between these unseen garments, which suggests that further manipulations are required to improve the classification prediction. Fig. \ref{fig:sequence} (\textit{right}) has suggested that the network holds a prediction as 'heavy' over a medium-weight garment. This is because heavy garments are sweaters and differ from the rest of the garments in terms of its materials. Therefore, our network can classify heavy garments but has a low-performance accuracy for shirts and pants. 

As opposed to shapes, weights are a more implicit physical property which are more difficult to be generalised. Nevertheless, the overall performance of the network ($48\%$ for shapes and $60\%$ for weights) suggests that our continuous perception hypothesis holds for garments with shapes such as pants, sweaters, towels, and t-shirts and with weights such as light and heavy, suggesting that further interactions with garments such as in \cite{5980336,7487399} are required to improve the overall classification performance. We must note that the overall shape classification performance while validating our network is approximately $90\%$; suggesting that the network can successfully predict known garment's shapes based on its dynamic properties.

\section{\uppercase{CONCLUSIONS}}

From the ablation studies we have conducted, depth images have a better performance over RGB images because depth captures the garment topology properties of garments. That is, our network was able to learn dynamic changes of the garments and make predictions on unseen garments since depth images have a prediction accuracy of $48\%$ and $60\%$ while classifying shapes and weights, accordingly. We also show that continuous perception improves classification accuracy. That is, weight classification, which is an indicator of garment physical properties, observes an increase in accuracy from 48.3\% to 60\% under a continous perception paradigm. This means that our network can learn physical properties from continuous perception. However, we observed an increase of around 1\% (from 47.6\% to 48\%) while continuously classifying garment's shape. The marginal improvement while continuously classifying shape indicates that further manipulations, such as flattening \cite{sun2015accurate} and unfolding \cite{doumanoglou2016folding} are required to bring a unknown garment to a state that can be recognised by a robot. That is, the ability to predict dynamic information of a piece of an unknown garment (or other deformable objects) facilitates robots' efficiency to manipulate it by ensuring how the garment will deform \cite{jimenez2020perception,ganapathi2020learning}. Therefore, an understanding of the dynamics of garments and other deformable objects can allow robots to accomplish grasping and manipulation tasks with higher dexterity

From the results, we can also observe that there exist incorrect classifications of unseen shirts because of their similarity in their materials. Therefore, we propose to experiment on how to improve prediction accuracy on garments with similar materials and structures by allowing a robot to interact with garments as proposed in \cite{7487399}. We also envisage that it can be possible to learn the dynamic physical properties (stiffness) of real garments from training a 'physical-similarity network' (PhysNet) \cite{runia2020cloth} on simulated garment models.

\bibliographystyle{apalike}
{\small
\bibliography{references}}

\begin{thebibliography}{}

\bibitem[Bhat et~al., 2003]{bhat2003estimating}
Bhat, K.~S., Twigg, C.~D., Hodgins, J.~K., Khosla, P.~K., Popovi{\'c}, Z., and
  Seitz, S.~M. (2003).
\newblock Estimating cloth simulation parameters from video.
\newblock In {\em Proceedings of the 2003 ACM SIGGRAPH/Eurographics symposium
  on Computer animation}, pages 37--51. Eurographics Association.

\bibitem[{Bohg} et~al., 2017]{8007233}
{Bohg}, J., {Hausman}, K., {Sankaran}, B., {Brock}, O., {Kragic}, D., {Schaal},
  S., and {Sukhatme}, G.~S. (2017).
\newblock Interactive perception: Leveraging action in perception and
  perception in action.
\newblock {\em IEEE Transactions on Robotics}, 33(6):1273--1291.

\bibitem[Davis et~al., 2015]{davis2015visual}
Davis, A., Bouman, K.~L., Chen, J.~G., Rubinstein, M., Durand, F., and Freeman,
  W.~T. (2015).
\newblock Visual vibrometry: Estimating material properties from small motion
  in video.
\newblock In {\em Proceedings of the ieee conference on computer vision and
  pattern recognition}, pages 5335--5343.

\bibitem[Doumanoglou et~al., 2016]{doumanoglou2016folding}
Doumanoglou, A., Stria, J., Peleka, G., Mariolis, I., Petrik, V., Kargakos, A.,
  Wagner, L., Hlav{\'a}{\v{c}}, V., Kim, T.-K., and Malassiotis, S. (2016).
\newblock Folding clothes autonomously: A complete pipeline.
\newblock {\em IEEE Transactions on Robotics}, 32(6):1461--1478.

\bibitem[Ganapathi et~al., 2020]{ganapathi2020learning}
Ganapathi, A., Sundaresan, P., Thananjeyan, B., Balakrishna, A., Seita, D.,
  Grannen, J., Hwang, M., Hoque, R., Gonzalez, J.~E., Jamali, N., Yamane, K.,
  Iba, S., and Goldberg, K. (2020).
\newblock Learning to smooth and fold real fabric using dense object
  descriptors trained on synthetic color images.

\bibitem[He et~al., 2016]{he2016deep}
He, K., Zhang, X., Ren, S., and Sun, J. (2016).
\newblock Deep residual learning for image recognition.
\newblock In {\em Proceedings of the IEEE conference on computer vision and
  pattern recognition}, pages 770--778.

\bibitem[Jim{\'e}nez and Torras, 2020]{jimenez2020perception}
Jim{\'e}nez, P. and Torras, C. (2020).
\newblock Perception of cloth in assistive robotic manipulation tasks.
\newblock {\em Natural Computing}, pages 1--23.

\bibitem[Krizhevsky et~al., 2012]{krizhevsky2012imagenet}
Krizhevsky, A., Sutskever, I., and Hinton, G.~E. (2012).
\newblock Imagenet classification with deep convolutional neural networks.
\newblock In {\em Advances in neural information processing systems}, pages
  1097--1105.

\bibitem[Li et~al., 2020]{li2020visual}
Li, Y., Lin, T., Yi, K., Bear, D., Yamins, D. L.~K., Wu, J., Tenenbaum, J.~B.,
  and Torralba, A. (2020).
\newblock Visual grounding of learned physical models.

\bibitem[Li et~al., 2018]{li2018learning}
Li, Y., Wu, J., Tedrake, R., Tenenbaum, J.~B., and Torralba, A. (2018).
\newblock Learning particle dynamics for manipulating rigid bodies, deformable
  objects, and fluids.
\newblock {\em arXiv preprint arXiv:1810.01566}.

\bibitem[Mariolis et~al., 2015]{mariolis2015pose}
Mariolis, I., Peleka, G., Kargakos, A., and Malassiotis, S. (2015).
\newblock Pose and category recognition of highly deformable objects using deep
  learning.
\newblock In {\em 2015 International Conference on Advanced Robotics (ICAR)},
  pages 655--662. IEEE.

\bibitem[Martínez et~al., 2019]{MARTINEZ2019220}
Martínez, L., del Solar, J.~R., Sun, L., Siebert, J.~P., and Aragon-Camarasa,
  G. (2019).
\newblock Continuous perception for deformable objects understanding.
\newblock {\em Robotics and Autonomous Systems}, 118:220 -- 230.

\bibitem[Runia et~al., 2020]{runia2020cloth}
Runia, T.~F., Gavrilyuk, K., Snoek, C.~G., and Smeulders, A.~W. (2020).
\newblock Cloth in the wind: A case study of physical measurement through
  simulation.
\newblock {\em arXiv preprint arXiv:2003.05065}.

\bibitem[Sengupta et~al., 2020]{sengupta2020simultaneous}
Sengupta, A., Lagneau, R., Krupa, A., Marchand, E., and Marchal, M. (2020).
\newblock Simultaneous tracking and elasticity parameter estimation of
  deformable objects.
\newblock In {\em IEEE Int. Conf. on Robotics and Automation, ICRA'20}.

\bibitem[Simonyan and Zisserman, 2014]{simonyan2014very}
Simonyan, K. and Zisserman, A. (2014).
\newblock Very deep convolutional networks for large-scale image recognition.
\newblock {\em arXiv preprint arXiv:1409.1556}.

\bibitem[Sun et~al., 2015]{sun2015accurate}
Sun, L., Aragon-Camarasa, G., Rogers, S., and Siebert, J.~P. (2015).
\newblock Accurate garment surface analysis using an active stereo robot head
  with application to dual-arm flattening.
\newblock In {\em 2015 IEEE international conference on robotics and automation
  (ICRA)}, pages 185--192. IEEE.

\bibitem[Sun et~al., 2017]{sun2017singleshot}
Sun, L., Aragon-Camarasa, G., Rogers, S., Stolkin, R., and Siebert, J.~P.
  (2017).
\newblock Single-shot clothing category recognition in free-configurations with
  application to autonomous clothes sorting.

\bibitem[{Sun} et~al., 2016]{7487399}
{Sun}, L., {Rogers}, S., {Aragon-Camarasa}, G., and {Siebert}, J.~P. (2016).
\newblock Recognising the clothing categories from free-configuration using
  gaussian-process-based interactive perception.
\newblock In {\em 2016 IEEE International Conference on Robotics and Automation
  (ICRA)}, pages 2464--2470.

\bibitem[Tanaka et~al., 2019]{tanaka2019learning}
Tanaka, D., Tsuda, S., and Yamazaki, K. (2019).
\newblock A learning method of dual-arm manipulation for cloth folding using
  physics simulator.
\newblock In {\em 2019 IEEE International Conference on Mechatronics and
  Automation (ICMA)}, pages 756--762. IEEE.

\bibitem[{Willimon} et~al., 2011]{5980336}
{Willimon}, B., {Birchfield}, S., and {Walker}, I. (2011).
\newblock Classification of clothing using interactive perception.
\newblock In {\em 2011 IEEE International Conference on Robotics and
  Automation}, pages 1862--1868.

\bibitem[Yang et~al., 2017]{yang2017learning}
Yang, S., Liang, J., and Lin, M.~C. (2017).
\newblock Learning-based cloth material recovery from video.
\newblock In {\em Proceedings of the IEEE International Conference on Computer
  Vision}, pages 4383--4393.

\end{thebibliography}
\end{document}